\documentclass[runningheads]{llncs}
\usepackage{graphicx}

\usepackage{amsmath}
\usepackage{amssymb}
\usepackage[linesnumbered,ruled,vlined]{algorithm2e}
\usepackage{multirow}

\usepackage{url}
\usepackage{subfig}

\usepackage[breaklinks,colorlinks,linkcolor={blue},citecolor={blue},urlcolor={blue}]{hyperref}

\usepackage[capitalize]{cleveref}

\usepackage{bbding}
\begin{document}

\newcommand{\x}{\mathbf{x}}
\newcommand{\y}{\mathbf{y}}
\newcommand{\g}{\mathbf{g}}
\newcommand{\w}{\mathbf{w}}
\newcommand{\z}{\mathbf{z}}

%
\title{Improving Handwritten OCR with Training Samples Generated by Glyph Conditional Denoising Diffusion Probabilistic Model}
%
%
\newcommand*\samethanks[1][\value{footnote}]{\footnotemark[#1]}

\author{
Haisong Ding\inst{1} 
\thanks{Corresponding author.}
\and 
Bozhi Luan\inst{2} \thanks{This work was done when Bozhi Luan and Dongnan Gui worked as interns in MMI Group, Microsoft
Research Asia, Beijing, China.} 
\and 
Dongnan Gui\inst{2} \samethanks
\and
Kai Chen\inst{1} \samethanks[0]{*}
\and
Qiang Huo\inst{1} 
}
%
%
\authorrunning{Ding, et al.}

\institute{
Microsoft Research Asia, Beijing, China \\
\email{dinghs11@mail.ustc.edu.cn},
\email{chenkai.cn@hotmail.com},
\email{qianghuo@microsoft.com}
\and
University of Science and Technology of China \\
\email{\{lbz0075,gdn2001\}@mail.ustc.edu.cn} 
}

\titlerunning{Improved Handwritten OCR with GC-DDPM Generated Training Samples}

\maketitle              

\setcounter{footnote}{0}

\begin{abstract}

Constructing a highly accurate handwritten OCR system requires large amounts of representative training data, which is both time-consuming and expensive to collect. To mitigate the issue, we propose a denoising diffusion probabilistic model (DDPM) to generate training samples. This model conditions on a printed glyph image and creates mappings between printed characters and handwritten images, thus enabling the generation of photo-realistic handwritten samples with diverse styles and unseen text contents. However, the text contents in synthetic images are not always consistent with the glyph conditional images, leading to unreliable labels of synthetic samples. To address this issue, we further propose a progressive data filtering strategy to add those samples with a high confidence of correctness to the training set. Experimental results on IAM benchmark task show that OCR model trained with augmented DDPM-synthesized training samples can achieve about $45\%$ relative word error rate reduction compared with the one trained on real data only.


\end{abstract}

\section{Introduction}

In recent years, researchers in handwritten optical character recognition (OCR) area are continuously making progress by leveraging advanced model architectures (e.g., \cite{Etter-ICDAR-2019,HWR-ICDAR-2021-1,HWR-ICDAR-2021-2,Rethink-Google-Arxiv-2021,TROCR-2021,HWR-ICFHR-2022,Wick-DAS-2022}). However, it is still a challenging problem, due to the cursive nature of handwritten strokes and diverse writing styles. 
To achieve excellent recognition accuracy for a handwritten OCR system, large amounts of labeled handwritten images are required. The handwritten image dataset should be representative enough to cover diverse writing styles and text contents. Obviously, collecting and labeling such a dataset are both time-consuming and expensive, and existing training data are limited in terms of style and content coverage. For example, as observed in \cite{SLOGAN-2022}, in the popular IAM dataset \cite{IAM}, a limited number of samples are collected for each writer, and some words are only written by a few writers. To handle this content-style data representation issue, one potential solution is to train a handwritten image generator to synthesize training samples for handwritten OCR. For any given text and a writer style, the generator should be able to synthesize photo-realistic handwritten images that can match the content of the input text and style of the writer.


In the past several years, generative adversarial network (GAN) \cite{GAN,cGAN,StyleGAN} based handwritten image generation methods have achieved promising results. Most of GAN-based handwritten image generation approaches adopt a text-to-image framework \cite{Alonso-ICDAR-2019,GANwriting-2020,ScrabbleGAN-2020,Davis-GAN-2020,HiGAN-2021,HTransformers-2021,JokerGAN-2021,Kang-GAN-PAMI-2022,GAN-ICPR-2022}. Given an input text and a writer embedding, it is able to generate a photo-realistic handwritten image that matches the content of the input text and style of the writer. However, using text as input is not sufficiently flexible to embed various contents such as adjacent character interval and character arrangements \cite{SLOGAN-2022}. By rendering text to a printed glyph image, SLOGAN \cite{SLOGAN-2022} proposed to use an image-to-image framework for handwritten image generation. It is able to generate flexible contents by rearranging characters on the input glyph image. It is noted that these GAN-based approaches all rely on guidance from an external handwritten recognizer trained on real data, which implies that the ability of GANs is limited to directly learn the mapping from texts or printed glyph images to handwritten images without external recognition model guidance.

Recently, denoising diffusion probabilistic models (DDPMs) \cite{DDPM-2020,DDPM-2015} achieve superior performances compared with other generation techniques on image generation tasks, including text-to-image generation \cite{BeatsGAN-2021,SDE-2021,DALLE2-2022,IMAGEN-2022} and image-to-image generation \cite{PITI-2022,RePaint-2022}. For handwritten image generation task, \cite{DDPM-HTG} investigated a text-to-image DDPM for online handwritten generation and achieved promising results. \cite{GC-DDPM} proposed a writer dependent glyph conditional DDPM (GC-DDPM) for offline handwritten Chinese character generation. GC-DDPM conditions on a printed glyph image and creates mappings between printed Chinese characters and handwritten images. Training from samples of a small Chinese character set, the GC-DDPM is capable of generating photo-realistic handwritten samples of unseen Chinese character categories. In \cite{GC-DDPM}, the DDPM is trained on a large-scale handwritten Chinese character database, where the number of training samples for each writer is relatively sufficient. In this paper, we investigate GC-DDPM on the offline English handwritten image generation task. We conduct experiments on the popular IAM dataset \cite{IAM} with limited training samples and content-style representation coverage. We find that even with limited training data, the GC-DDPM is still able to generate photo-realistic handwritten images. 

Since no explicit recognition model guidance is adopted in GC-DDPM, during sampling, the model can generate noisy samples where the synthesized images do not match the text contents in glyph conditional images. Directly adding these samples to the training set for OCR can degrade the recognition performance. To address this problem, inspired by the self-training framework in automatic speech recognition \cite{ST-ICASSP-2020,ST-INTERSPEECH-2020} and OCR \cite{ST-ICPR-2022,ST-ICFHR-2022}, we propose a progressive data filtering strategy to add samples with a high confidence of correctness to the training set. 
Experimental results on IAM benchmark task show that the performance of the OCR model can be significantly improved when trained with augmented DDPM-synthesized samples.

The remainder of this paper is organized as follows. In \cref{sec:related_work}, we briefly review related works. In \cref{sec:approach}, we adopt the GC-DDPM approach in \cite{GC-DDPM} for offline English handwritten image generation task and introduce the progressive data filtering strategy. Experimental results are presented in \cref{sec:exp}. Finally, we conclude this study in \cref{sec:conclude}.

\section{Related Works} \label{sec:related_work}

\subsection{GAN-based handwritten image generation approaches}

Handwritten image generation aims to synthesize offline handwritten images given input texts. Most of the previous approaches directly use a text-to-image framework based on GANs. For example, \cite{Alonso-ICDAR-2019} proposed a GAN-based handwritten word image generator with additional guidance from an external handwritten recognizer trained on real data. \cite{GANwriting-2020} further proposed a handwritten generation model to synthesize handwritten word images that can match the given conditional writing styles. ScrabbleGAN \cite{ScrabbleGAN-2020} and HiGAN \cite{HiGAN-2021} used fully-convolutional generators, which can generate images of words and text lines with arbitrary lengths by making the image width proportional to the length of input text. In \cite{Davis-GAN-2020}, the image widths are also automatically learned based on the input text and style. The text-to-image handwritten image generation framework is further improved with advanced generation model architectures such as self-attention and deformable convolution layers \cite{HTransformers-2021,Kang-GAN-PAMI-2022,GAN-ICPR-2022}.

Since only text inputs are leveraged, the generation model needs to learn the mapping from text embedding to handwritten strokes, which is quite difficult. Besides pure text, JokerGAN \cite{JokerGAN-2021} proposed to leverage an additional text line clue about the existence of ``below the baseline'' and ``above the mean line'' characters to improve the generation model. By rendering text to a printed glyph image using a standard typeset font, the resulting glyph image obviously contains more information than text. Using the glyph image as input, SLOGAN \cite{SLOGAN-2022} proposed to use an image-to-image framework for handwritten image generation. It is able to generate flexible contents by changing the positions of characters and adjusting space interval between adjacent characters in glyph images. Besides using text or glyph image as input, \cite{SCGAN-ICFHR-2020} proposed to synthesize handwritten images from online handwritten samples based on StyleGAN \cite{StyleGAN}.

Many of the above-mentioned approaches leveraged synthesized handwritten images as training data to boost the performances of handwritten OCR systems, and achieved substantial improvements (e.g., \cite{SCGAN-ICFHR-2020,Kang-GAN-PAMI-2022,SLOGAN-2022}). 
In this paper, we investigate DDPM to synthesize handwritten images to augment the training data for handwritten OCR.

\subsection{Diffusion model}

DDPMs \cite{DDPM-2015,DDPM-2020} have been extremely popular in image generation tasks. DDPM defines a Markov chain of $T$ diffusion steps. In a forward diffusion process, it slowly corrupts data by adding random noises, then a reverse diffusion process is learned to recreate data from Gaussian noise. It is shown in \cite{BeatsGAN-2021} that DDPMs can outperform GANs on image synthesis. In \cite{GLIDE-2022}, DDPMs for text-to-image synthesis are explored. The model is able to generate photo-realistic images that match the content of conditional text with the help of a classifier-free guidance \cite{Classfier-Free-2021}. Furthermore, in \cite{DALLE2-2022,IMAGEN-2022}, DDPMs have demonstrated powerful capabilities to generate high-quality images given input texts. DDPMs are also successfully applied to other tasks such as image-to-image generation (e.g., \cite{PITI-2022}). \cite{DDPM-HTG} investigated DDPM for online handwritten generation and achieved promising results. \cite{GC-DDPM} proposed a GC-DDPM for offline handwritten Chinese character generation. The GC-DDPM conditions on a printed glyph image and learns the mappings between printed Chinese character images and handwritten ones. It is able to generate photo-realistic handwritten images of unseen Chinese character categories. In this paper, we adopt GC-DDPM to generate offline English handwritten images.

\section{Our Approach} \label{sec:approach}

\subsection{GC-DDPM for handwritten image generation}

\begin{figure}[!t]
    \centering
    \subfloat[The Markov chain of forward diffusion process $q(\x _{t} | \x _{t-1})$ and the learned reverse diffusion process $p_{\boldsymbol{\theta}}(\x_{t-1} | \x_{t})$ of DDPM.]{
    \label{fig:ddpm markov}
        \includegraphics[width=0.9\linewidth]{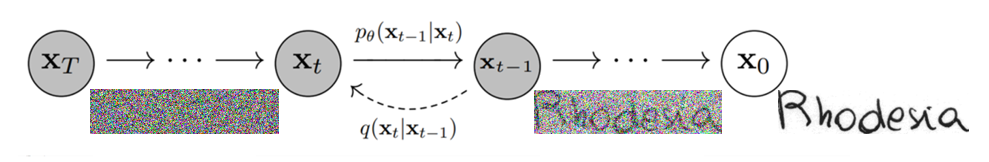}
    }
    \vspace{0mm}
     \subfloat[The architecture of the U-Net to estimate $\boldsymbol{\epsilon_{\theta}}(\x_t, \g, \w)$ and $\boldsymbol{\Sigma_{\theta}}(\x_t, \g, \w)$. Figures are adapted from \cite{GC-DDPM}.]{
    \includegraphics[width=0.9\linewidth]{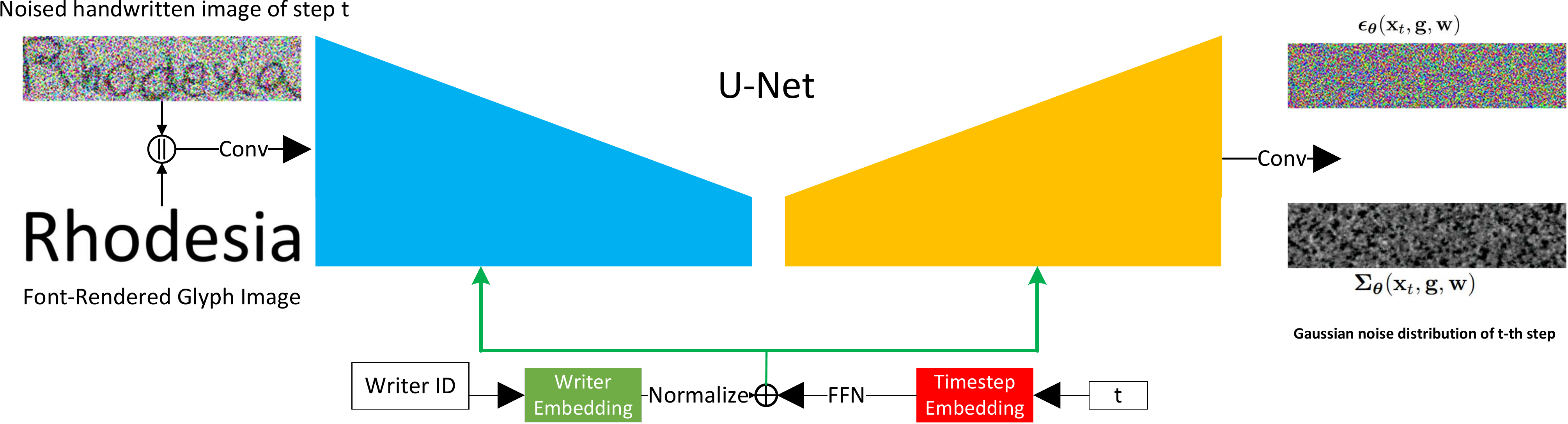}
    \label{fig:ddpm}
    }
    \caption{Illustration of GC-DDPM framework for handwritten image generation.
    }
    
\end{figure}

Given an input text and a writer ID (denoted as $\w$), we adopt a writer dependent GC-DDPM \cite{GC-DDPM} to generate photo-realistic handwritten images that match the content of the text and style of the writer. For each input text, we directly render it to a printed glyph image using a standard glyph font. It is more suitable to use a glyph image as input because it contains much more information about the shapes of individual characters than pure text. We denote the glyph image as $\g$.
As shown in Fig. \ref{fig:ddpm markov}, let $\x_0$ denote a data sampled from a real distribution, i.e., $\x_0 \sim q(\x)$ with a corresponding writer ID $\w$ and glyph image $\g$. In the forward diffusion process, small amounts of Gaussian noise are added to $\x$ in $T$ steps, producing a sequence $\{ \x_t \}_{t=1}^T$ calculated as follows:
\begin{equation} \label{eqn:ddpm_forward}
   q(\x _{t} | \x _{t-1}) =\mathcal{N}(\x _{t} ; \sqrt{1-\beta _{t}} \x _{t-1}, \beta _{t} \mathbf{I}), \quad \x _{t}= \sqrt{\alpha _t}\x _{t-1} + \sqrt{1-\alpha_t} \boldsymbol {\epsilon}_t \;,
\end{equation}
where $\boldsymbol {\epsilon}_t \sim \mathcal{N}(\mathbf{0}, \mathbf{I})$, $\beta_t \in (0, 1)$ and $\alpha_t = 1 - \beta_t$. It is easy to calculate that
\begin{equation}
   q(\x _{t} | \x _0) =\mathcal{N}(\x _{t} ; \sqrt{\bar{\alpha}_t} \x _{0}, (1 - \bar{\alpha}_t) \mathbf{I}), \quad \x _{t} = \sqrt{\bar{\alpha} _t}\x _{0} + \sqrt{1-\bar{\alpha}_t} \boldsymbol {\epsilon} \;,
\end{equation}
where $\boldsymbol {\epsilon} \sim \mathcal{N}(\mathbf{0}, \mathbf{I})$, $\bar{\alpha}_t = \prod_{i=1}^t \alpha_i$. When $T \rightarrow \infty$, $\bar{\alpha}_T \rightarrow 0$, and $\x _{T} \in \mathcal{N}(\mathbf{0}, \mathbf{I})$. A nice property of the forward diffusion process is that the reverse conditional probability is Gaussian when conditioned on $\x_0$:
\begin{equation}
    q(\x_{t-1} | \x_{t},\x_0)=\mathcal{N}(\x_{t-1};\tilde{\boldsymbol{\mu}}(\x_t,\x_0),\tilde{\beta_t}\mathbf{I}) \;,
\end{equation}
where
\begin{equation}
    \tilde{\boldsymbol{\mu}}(\x_t,\x_0) = \frac{1}{\sqrt{\alpha_t}}(\x_t - \frac{1-\alpha_t}{\sqrt{1-\bar{\alpha}_t}}\boldsymbol{\epsilon}_t), \quad
    \tilde{\beta_t}=\frac{1-\bar{\alpha}_{t-1}}{{1-\bar{\alpha}_t}}\cdot\beta_t \;.
\end{equation}
Moreover, the reverse process as shown in Fig. \ref{fig:ddpm markov} will also be a Gaussian when $\beta_t$ is sufficiently small. Therefore, we can learn a model $p_{ \boldsymbol{\theta} }$ to approximate the reverse process conditioned on $\g$ and $\w$:
\begin{equation} \label{eqn:ddpm}
    p_{\boldsymbol{\theta}}(\x_{t-1} | \x_{t},\g,\w)=\mathcal{N}(\x_{t-1};\boldsymbol{\mu_{\theta}}(\x_t,\g,\w),\boldsymbol{\Sigma_{\theta}}(\x_t,\g,\w)) \;.
\end{equation}
Following \cite{IDDPM-2021,GC-DDPM}, $\boldsymbol{\mu_{\theta}}(\x_t,\g,\w)$ and $\boldsymbol{\Sigma_{\theta}}(\x_t,\g,\w)$ are re-parameterized as
\begin{align} \label{eqn:re-pa}
\boldsymbol{\mu_{\theta}}(\x_t,\g,\w) &= \frac{1}{\sqrt{\alpha_t}} \left( \x_t - \frac{1-\alpha_t}{\sqrt{1-\bar{\alpha}_t}}\boldsymbol{\epsilon}_{\boldsymbol{\theta}}(\x_t, \g,\w) \right) \\
\boldsymbol{\Sigma_{\theta}}(\x_t,\g,\w) &= \exp\left(\boldsymbol{\nu}_{\theta}(\x_t, \g,\w)\log{\beta_t}
    +(1-\boldsymbol{\nu}_{\theta}(\x_t, \g,\w))\log{\tilde{\beta_t}} \right) \nonumber \;.
\end{align}

A neural network is trained to estimate $\boldsymbol{\epsilon_{\theta}}(\x_t, \g,\w)$ and $\boldsymbol{\Sigma}_{\theta}(\x_t, \g,\w)$. We use the same hybrid objective function as in \cite{IDDPM-2021}. After the reverse process is learned, conditioned on $\g$ and $\w$, we are able to draw samples $\x _0$ according to Eqn. (\ref{eqn:ddpm}), starting with a Gaussian noise $\x _T \sim \mathcal{N}(\mathbf{0}, \mathbf{I})$. 

We adopt the same U-Net architecture in GC-DDPM \cite{BeatsGAN-2021,GC-DDPM} in our task. As shown in Fig. \ref{fig:ddpm}, $\x_t$ and $\g$ are normalized to a fixed size. Then they are concatenated together and used as input to the U-Net. Time step $t$ is embedded with sinusoidal embedding, and then processed with a 2-layer feed-forward network (FFN). Writer information $\w$ is embedded with a learnable embedding, followed by L2-normalization: $\z = \w / \| \w \|_2$. Finally, they are added together and fed to layers in U-Net using a feature-wise linear modulation (FiLM) operator \cite{FiLM-2018}.

In DDPM, classifier-free guidance \cite{Classfier-Free-2021} is an effective approach to improve generation quality. Following \cite{GC-DDPM}, a content guidance scale $\gamma$ and a style content scale $\eta$ are used.
During sampling, $\boldsymbol{\epsilon}_{\theta}(\x_t, \g, \w)$ is directly replaced with
\begin{align} \label{eqn:cf_3}
    \boldsymbol{\tilde{\epsilon}}_{\theta}(\x_t, \g, \w) &= 
    \boldsymbol{\epsilon}_{\boldsymbol{\theta}}(\x_t, \g,\w)
    + \gamma \boldsymbol{\epsilon}_{\boldsymbol{\theta}}(\x_t, \g, \emptyset)  \\
    &+ \eta \boldsymbol{\epsilon}_{\boldsymbol{\theta}}(\x_t,\emptyset, \w)
    - (\gamma+\eta)\boldsymbol{\epsilon}_{\boldsymbol{\theta}}(\x_t \emptyset, \emptyset) \nonumber \;.
\end{align}
Here ${\boldsymbol{\epsilon_\theta}}(\x_t, \g, \emptyset)$, ${\boldsymbol{\epsilon_\theta}}(\x_t, \emptyset, \w)$ and ${\boldsymbol{\epsilon_\theta}}(\x_t, \emptyset, \emptyset)$ are trained together with ${\boldsymbol{\epsilon_\theta}}(\x_{t}, \g, \w)$ using the same U-Net, where $\w$, $\g$ or both are replaced with a special token $\emptyset$.

Works in \cite{GC-DDPM} synthesize offline Chinese character images which are of fixed height and width. While in English handwritten image generation task, the widths of the generated images should be inferred from the text and writer style. To achieve this goal, we prepare handwritten images of words and short phrases to train GC-DDPM, where the maximum aspect ratio of images is set as 8. First, images are resized to a height of 64 while keeping aspect ratio. Then, images are padded to a width of 512 with black pixels on both left and right margins. The glyph images are also processed to the size of $64\times 512$ using the same procedure. In experiments, we find that during sampling, GC-DDPM will learn the width of black margins based on input text and writer style. It will generate handwritten images with clear black margins robustly. Therefore, to get the final handwritten sample, we use a simple image processing method to remove the padded black margins.

\subsection{Progressive data filtering strategy}

By conditioning on glyph images and writer IDs, we expect GC-DDPM to learn the mapping from glyph images to handwritten ones and generate high-quality training data to improve handwritten OCR systems. However, we notice that the generated images are not always consistent with the glyph conditional images. Adding these data to the training set would degrade the performance and robustness of the handwritten OCR system. To alleviate this problem, we propose a progressive data filtering strategy to remove these noisy samples.

In text-to-image generation tasks, a dot product score between text and image embeddings is used as a metric to select generated samples to improve generation quality \cite{DALLE2-2022}. In the self-training framework, for each unlabeled sample with a pseudo label, a confidence score is calculated. Only data with high confidence scores are added to the training set \cite{ST-ICASSP-2020,ST-INTERSPEECH-2020,ST-ICPR-2022,ST-ICFHR-2022}. Inspired by these works, we design a metric to estimate the ``confidence of correctness'' for each generated handwritten image. Then samples with a high confidence of correctness are added to the training set progressively.

Let $\mathbf{R}=\{ (\boldsymbol{x}_i, \boldsymbol{y}_i) \} $ denote a real dataset with handwritten image $\boldsymbol{x}_i$ and ground truth label $ \boldsymbol{y}_i$. Let $\mathbf{S}=\{ (\tilde{\boldsymbol{x}}_j, \tilde{\boldsymbol{y}}_j) \} $ denote DDPM-generated synthetic dataset with handwritten image $\tilde{\boldsymbol{x}}_j$ and corresponding conditional text $\tilde{\boldsymbol{y}}_j$. First, we train an initial OCR model $\boldsymbol{M}$ using $\mathbf{R}$ only. Then the trained model is used to calculate the confidence of correctness score for each $\tilde{\boldsymbol{x}}_j$ in $\mathbf{S}$ as follows:
\begin{equation} \label{eqn:confidence}
    c(\tilde{\boldsymbol{x}}_j, \tilde{\boldsymbol{y}}_j ; \boldsymbol{M}) = \frac{ \mathcal{L}(\tilde{\boldsymbol{x}}_j, \hat{\boldsymbol{y}}_j; \boldsymbol{M}) }{ \mathcal{L} (\tilde{\boldsymbol{x}}_j, \tilde{\boldsymbol{y}}_j; \boldsymbol{M}) } \;,
\end{equation}
where $\mathcal{L}(\boldsymbol{x}, \boldsymbol{y}; \boldsymbol{M}) = -\log p(\boldsymbol{y} | \boldsymbol{x}; \boldsymbol{M})$ is the negative log posterior probability calculated using recognizer $\boldsymbol{M}$, and $\hat{\boldsymbol{y}}_j$ is the decoding result of $\tilde{\boldsymbol{x}}_j$ using $\boldsymbol{M}$. Obviously, if $\tilde{\boldsymbol{y}}_j = \hat{\boldsymbol{y}}_j$, the score equals to 1, meaning that the recognizer's prediction is consistent with the conditional text. Then the confidence of $\tilde{\boldsymbol{x}}$ matching $\tilde{\boldsymbol{y}}$ is high. If $\mathcal{L}(\tilde{\boldsymbol{x}}_j, \hat{\boldsymbol{y}}_j; \boldsymbol{M}) \ll \mathcal{L}(\tilde{\boldsymbol{x}}_j, \tilde{\boldsymbol{y}}_j; \boldsymbol{M})$, the score is close to 0, and the confidence of $\tilde{\boldsymbol{x}}$ matching $\tilde{\boldsymbol{y}}$ is low. In practice, a threshold $\tau \in (0, 1]$ is used, and $(\tilde{\boldsymbol{x}}_j, \tilde{\boldsymbol{y}}_j)$ with $c(\tilde{\boldsymbol{x}}_j, \tilde{\boldsymbol{y}}_j ; \boldsymbol{M}) \geq \tau$ is included in a selected set $\mathbf{S}'$. Then, a new OCR model can be trained with $\mathbf{R} \cup \mathbf{S}'$. After that, the scores can be re-calculated using the new model. This process can be repeated for multiple rounds until the performance of the OCR model does not improve further. The whole progressive data filtering strategy is summarized in Algorithm \ref{alg:data_filt}.

\begin{algorithm}[t]
    \caption{Progressive data filtering strategy}\label{alg:data_filt}

    \SetKwInput{KwInput}{Input}
    \KwInput{Real data $\mathbf{R}=\{ (\boldsymbol{x}_i, \boldsymbol{y}_i) \} $, synthetic data $\mathbf{S}=\{ (\tilde{\boldsymbol{x}}_j, \tilde{\boldsymbol{y}}_j) \}$, initial selected synthetic data $\mathbf{S}'=\{\}$, number of progressive data filtering rounds $N$, data filtering threshold $\tau$ }
    Train model $\boldsymbol{M}$ using $\mathbf{R}$\;
    \For{$n \gets 1$ to $N$ } {
        $\mathbf{S}' = \left\{ (\tilde{\boldsymbol{x}}_j, \tilde{\boldsymbol{y}}_j) \in \mathbf{S} ~|~ c(\tilde{\boldsymbol{x}}_j, \tilde{\boldsymbol{y}}_j ; \boldsymbol{M}) \geq \tau  \right\} $\;
        Train model $\boldsymbol{M}$ using $\mathbf{R} \cup \mathbf{S}'$ starting from random weight initialization\;
    }
    \KwRet{ $\boldsymbol{M}$, $\mathbf{S}'$}\;
\end{algorithm}

\section{Experiments} \label{sec:exp}

\subsection{Experimental setup}

We conduct experiments on the IAM dataset \cite{IAM}. It contains 13,353 isolated text line images and 115,320 word images written by 657 different writers. We use the RWTH Aachen partition as in \cite{Kang-ICFHR-2020} in experiments. Following \cite{BeatsGAN-2021}, diffusion step number $T$ is set as 1,000 with a linear noise schedule. During training, $\w$ and $\g$ are randomly set to $\emptyset$ with probability 10\%, independently. When $\g=\emptyset$, a blank glyph image will be used; when $\w=\emptyset$, a special embedding will be used. During sampling, we use DDIM \cite{DDIM-2021} sampling method with 50 steps to save sampling time. As for the handwritten OCR system, the same CTC-based \cite{CTC-ICML-2006} model in \cite{SLOGAN-2022} is used without leveraging external language models, and $\hat{\boldsymbol{y}}_j$ in Eqn. (\ref{eqn:confidence}) is decoded with the best path decoding algorithm. The total number of parameters of the OCR model is 14M. We mainly conduct experiments on the IAM word benchmark, except in \cref{sec:exp-line} where we conduct experiments on IAM text line benchmark. A word error rate (WER) of $19.47\%$ and a character error rate (CER) of $7.27\%$ is achieved when trained on real IAM word data only, which is similar to the baseline result ($19.12\%$ WER and $7.39\%$ CER) presented in \cite{SLOGAN-2022}.

\subsection{Effect of classifier-free guidance scales in GC-DDPM}

\begin{figure}[t]
    \centering
    \includegraphics[width=0.8\linewidth]{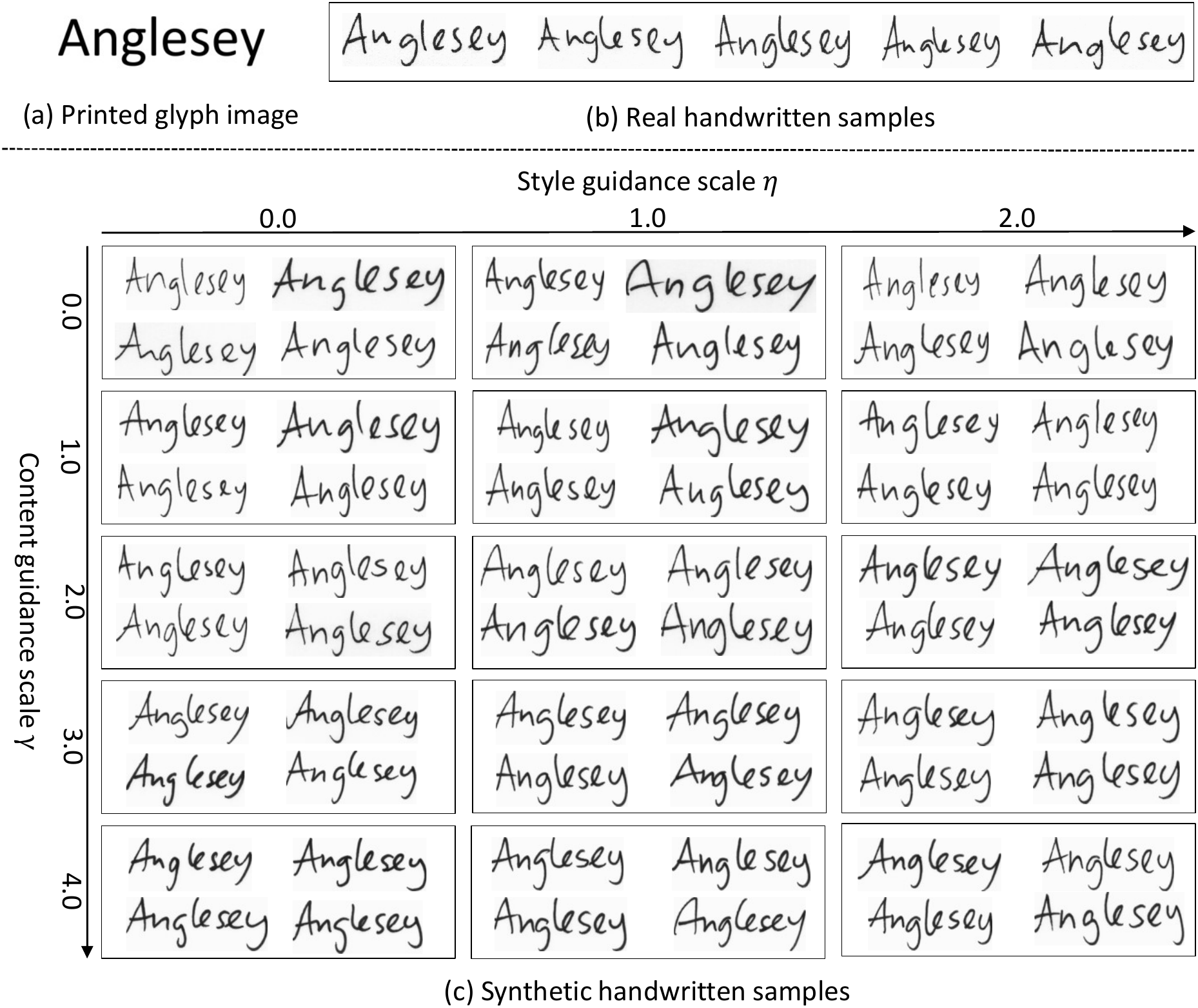}
    \caption{Real samples of words ``Anglesey'' written by writer 333 and synthetic handwritten images generated with different guidance scales. Glyph conditional image is shown in (a).}
    \label{fig:samples-scales}
\end{figure}

\begin{table}[t]
    \centering
    \caption{WER and CER of handwritten OCR models on IAM word testing set trained with synthetic dataset generated with different guidance scales.}
    \label{table:wer-cer-scales}
    \begin{tabular}{|c|c|c|c|}
        \hline
        Style scale $\eta$  & Content scale $\gamma$       & WER (\%) & CER (\%)  \\
        \hline \hline
        \multirow{3}{*}{0.0}    & 0.0 & \textbf{20.17} & \textbf{7.50} \\ \cline{2-4}
                                & 0.5 & 21.06 & 7.94 \\ \cline{2-4}
                                & 1.0 & 21.93 & 8.23 \\ \hline
        \multirow{3}{*}{0.5}    & 0.0 & 20.35 & 7.52 \\ \cline{2-4}
                                & 0.5 & 21.14 & 7.91 \\ \cline{2-4}
                                & 1.0 & 21.67 & 8.14 \\ \hline
        \multirow{3}{*}{1.0}    & 0.0 & 20.40 & 7.59 \\ \cline{2-4}
                                & 0.5 & 20.41 & 7.56 \\ \cline{2-4}
                                & 1.0 & 21.25 & 7.91 \\ \hline \hline
        \multicolumn{2}{|c|}{IAM training set} & 19.47 & 7.27 \\ \hline
    \end{tabular}
\end{table}

Works in \cite{Classfier-Free-2021,GC-DDPM} show that the classifier guidance scale is able to control the trade-off between the quality and diversity of generated samples. Fig. \ref{fig:samples-scales} (b) shows real samples of ``Anglesey'' written by writer 333. Clearly, the style and position of individual characters vary each time the same writer writes them. Fig. \ref{fig:samples-scales} (c) visualizes synthetic samples generated with different guidance scales. With higher content guidance scales, the generated samples become less diverse, which is consistent with the observation in \cite{GC-DDPM}. For example, in real samples, the ``le'' in ``Anglesey'' is either separately written, or consecutively written with a single stroke. Synthetic samples with lower content guidance scales successfully capture both writing variants. Whereas samples with higher content guidance scales only capture the consecutively written one. We also observe that the variance in generated image widths becomes smaller when sampled with higher content guidance scales. As for the style guidance scales, since the writer ID is already a distinctive guidance, the sampling qualities with different scales are similar.

To evaluate the behavior of guidance scales in generating training data for handwritten OCR, we try $\gamma, \eta \in \{0.0,~0.5,~1.0\}$ and synthesize the whole IAM word training set using the exact word corpus and writer IDs. The number of synthetic images equals to the number of images in the IAM training set. Then, we use synthetic data only to train an OCR model and evaluate its recognition performance on the real IAM word testing set. As shown in Table \ref{table:wer-cer-scales}, with the same $\eta$, WER increases as $\gamma$ becomes higher. This shows that the diversity of generated images is important when synthesizing training data for OCR. The best recognition performance is achieved with both guidance scales set as 0. The best WER is only absolute $0.7\%$ worse than that trained on real dataset, which demonstrates the high quality of DDPM-generated handwritten images. Based on these observations, we set $\gamma=0.0,~\eta=0.0$ in the following experiments.

\begin{figure}[t]
    \centering
    \includegraphics[width=0.8\linewidth]{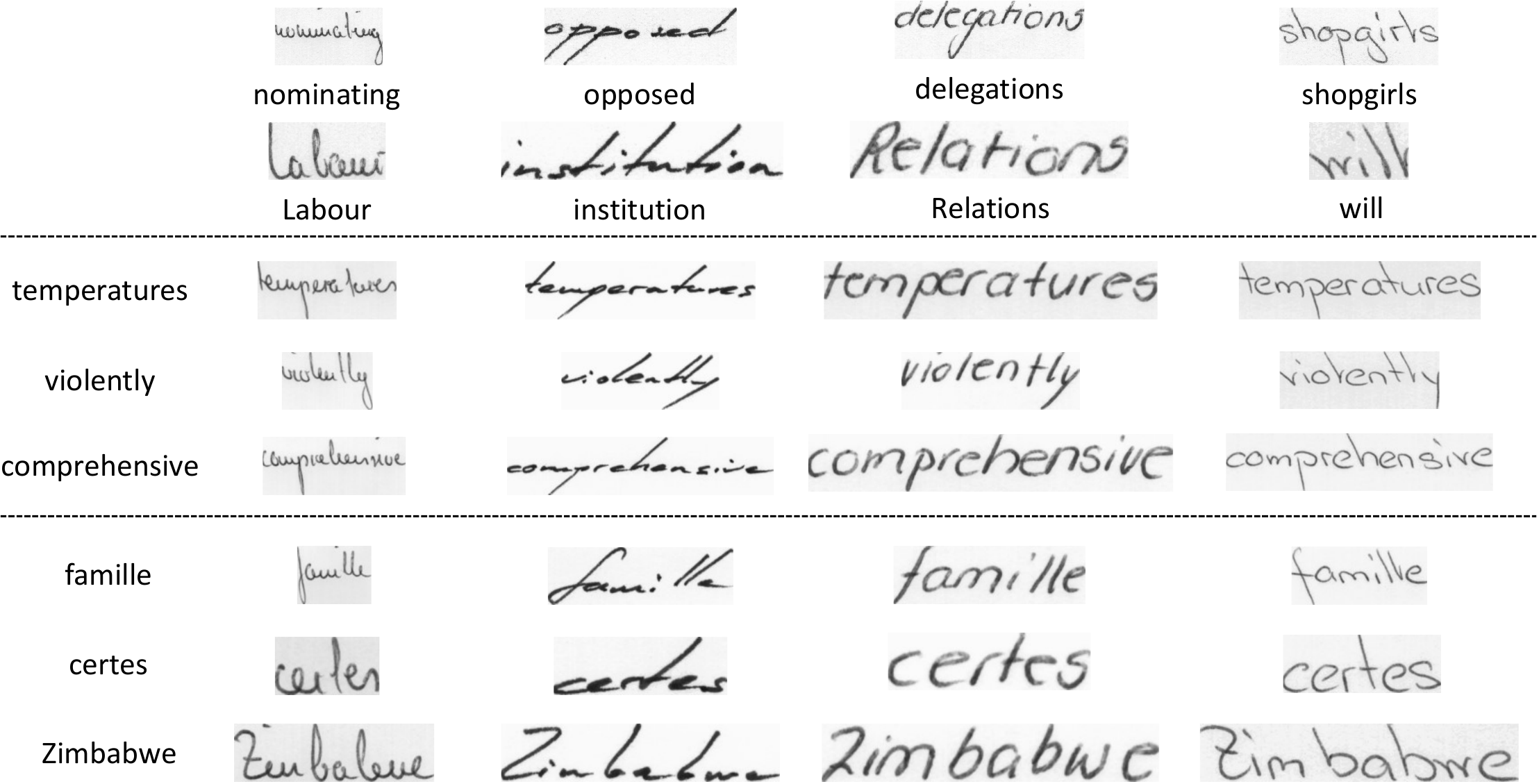}
    \caption{Synthetic handwritten images conditioned by different writer IDs with different words. Top: real samples with corresponding words from IAM writers 001, 002, 023 and 027, respectively. Middle: synthetic samples of words that are seen in IAM training set. Bottom: synthetic samples of out-of-vocabulary words. }
  \label{fig:samples-words}
\end{figure}

In Fig. \ref{fig:samples-words}, we show generated samples conditioned by different writer IDs with input words that are seen in IAM training set and out-of-vocabulary words. It is clear that GC-DDPM is able to synthesize these words while mimicing the writing styles (e.g., cursive, slant, stroke pattern) of the conditional writers. It is noted that although ``Z'' only appears four times in the training set, the GC-DDPM still can generate high quality handwritten images.

\subsection{Augment training set with synthetic images for OCR}

\begin{figure}[t]
    \centering
    \includegraphics[width=0.8\linewidth]{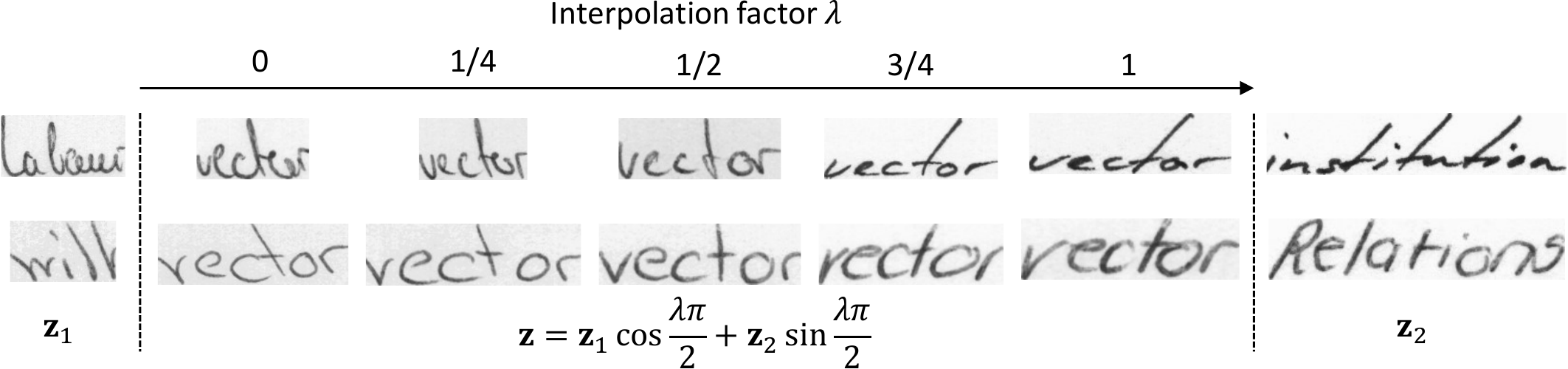}
    \caption{Synthetic handwritten images of word ``vector'' generated using writer style interpolations between $\z_1$ and $\z_2$.}
		\label{fig:samples-interp}
\end{figure}

\begin{table}[t]
    \centering
    \caption{WER and CER of handwritten OCR models on IAM word testing set trained with different training sets.}
    \label{table:wer-cer-comp}
    \begin{tabular}{|l|c|c|}
        \hline
        {Training set}           & WER(\%) & CER(\%)  \\
        \hline \hline
        IAM training set                & 19.47 & 7.27 \\ \hline \hline
        ~~~~+ Synth-IAM-Words       & 11.57 & 3.88 \\ \hline
        ~~~~+ Synth-EN-Words        & 14.78 & 5.14 \\ \hline
        ~~~~+ Synth-EN-Words-WI     & 14.83 & 5.18 \\ \hline
    \end{tabular}
\end{table}

Next, we use synthetic handwritten images to boost handwritten OCR performance. To evaluate the quality of generated samples of seen/unseen words, three sets of synthetic handwritten images are generated in our experiments:
\begin{itemize}
    
    \item[$\bullet$] Synth-IAM-Words: Since the IAM training set is insufficient in terms of content and style coverage, we use GC-DDPM to generate handwritten images for each writer. For each of the 442 writers observed in the training set, we synthesize a handwritten image for each and every word in the entire IAM word training corpus. As a result, the Synth-IAM-Words dataset is 442 times the size of the original training set in terms of the number of samples.
    
    \item[$\bullet$] Synth-EN-Words: Samples in Synth-IAM-Words only contain words that have been observed in training set. To investigate the quality of synthesized handwritten samples of unseen words, following \cite{ScrabbleGAN-2020,SLOGAN-2022}, an external ``English words'' \footnote{\url{https://github.com/dwyl/english-words/blob/master/words.txt}} corpus is used. It contains 466,550 unique words, 98.9\% of which are not observed in the IAM training set. To generate a diverse dataset, for each word, we synthesize 8 samples conditioning on 8 randomly selected writer IDs. As a result, Synth-EN-Words dataset contains about 3.7M samples.
    
    \item[$\bullet$] Synth-EN-Words-WI: The above two datasets are generated with trained writer IDs. The GC-DDPM is also able to generate unseen styles using writer style interpolations \cite{GC-DDPM}. To achieve this, given two normalized writer embedding $\z_1$ and $\z_2$, a new embedding $\z$ can be obtained with spherical interpolation \cite{DALLE2-2022}: $\z = \z_1 \cos \frac{\lambda \pi}{2} + \z_2 \sin \frac{\lambda \pi}{2}$ with interpolation factor $\lambda \in [0, 1]$. Fig. \ref{fig:samples-interp} shows handwritten samples generated using writer style interpolations. It is clear that as $\lambda$ increases from $0$ to $1$, the style of generated samples gradually shifts from $\z_1$ to $\z_2$. To evaluate the quality of synthesized handwritten samples with interpolated styles, for each word in ``English words'' corpus, we also synthesize 8 samples conditioning on 8 randomly calculated writer interpolations. We use $\lambda=1/2$. We name this dataset ``Synth-EN-Words-WI''. It also contains 3.7M samples. 
\end{itemize}

Table \ref{table:wer-cer-comp} lists the performances of handwritten OCR models on IAM word testing set trained with different training sets. It is clear that augmenting the training set with synthetic handwritten images can significantly boost the recognition performances. Specifically, a $40.6\%$ WER reduction (WERR) and a $46.6\%$ CER reduction (CERR) are achieved with Synth-IAM-Words dataset. It shows that the generated Synth-IAM-Words can successfully alleviate the insufficient content and style coverage problem in training set and achieves significant OCR performance improvements. Augmenting the training set with Synth-EN-Words and Synth-EN-Words-WI achieves similar recognition accuracy improvements (about $24\%$ WERR and $29\%$ CERR), which suggests that the synthesized qualities of Synth-EN-Words and Synth-EN-Words-WI are similar.

In our experiments, we construct synthetic datasets using two corpora, the IAM corpus and an external ``English words'' corpus. Words in IAM corpus are seen in the training of DDPM, whereas most of words in ``English words'' corpus are unseen. We find that the quality of synthesized data of words in ``English words'' is worse than that of words in IAM corpus. This shows that the synthesized data quality of unseen words is worse than that of seen words.

To further show the quality difference between Synth-IAM-Words and Synth-EN-Words/Synth-EN-Words-WI, we treat their conditional texts as ground truths and evaluate the WER using the OCR model trained on IAM training set. A $22\%$ WER is observed on Synth-IAM-Words, while the WERs on Synth-EN-Words and Synth-EN-Words-WI are $73\%$ and $72\%$, which are significantly higher. There are two potential reasons for this observation: (a) the generalization ability of the OCR model trained with IAM training set is limited when recognizing unseen words, and (b) the quality of DDPM-synthesized handwritten images of unseen words is worse than seen words. As a comparison, we evaluate the OCR model on a subset of IAM word testing set containing unseen words and achieve a $40.8\%$ WER. Therefore, both reasons contribute to the high WER. Based on these analyses, we conclude that the text contents in synthetic images are not always consistent with the glyph conditional images, leading to unreliable labels of synthetic data. Adding noisy data to the training set could degrade the performance of the OCR model. Next, we will leverage the progressive data filtering strategy to remove synthetic data with unreliable labels.

\subsection{Effect of progressive data filtering strategy}

\begin{table}[t]
    \centering
    \caption{WER and CER of handwritten OCR models on IAM word testing set trained on augmented synthetic datasets with progressive data filtering strategy.}
    \label{table:wer-cer-filt}
    \begin{tabular}{|l|c|c|c|c|c|c|c|}
        \hline
        \multirow{2}{*}{Synthetic dataset} & \multirow{2}{*}{$~N~$} & \multicolumn{2}{c|}{$\tau=1.0$} & \multicolumn{2}{c|}{$\tau=0.7$} & \multicolumn{2}{c|}{Use all samples} \\ \cline{3-8}
        & & WER(\%) & CER(\%) & WER(\%) & CER(\%) & WER(\%) & CER(\%) \\ \hline

        \multirow{3}{*}{Synth-IAM-Words} & 1 & 12.70 & 4.60 & 12.40 & 4.45 & \multirow{3}{*}{11.57} & \multirow{3}{*}{\textbf{3.88}} \\ \cline{2-6}
                                             & 2 & 11.84 & 4.23 & 11.73 & 4.16 & &  \\ \cline{2-6}
                                             & 3 & 11.77 & 4.22 & \textbf{11.54} & 4.14 & &  \\ 
        \hline
        \multirow{3}{*}{Synth-EN-Words}  & 1 & 15.33 & 5.38 & 14.53 & 5.09 & \multirow{3}{*}{14.78} & \multirow{3}{*}{5.14} \\ \cline{2-6}
                                             & 2 & 14.38 & 5.01 & 14.13 & 4.86 & &  \\ \cline{2-6}
                                             & 3 & 14.12 & 4.88 & \textbf{13.85} & \textbf{4.83} & &  \\ \hline
        
        \multirow{3}{*}{Synth-EN-Words-WI~}  & 1 & 15.18 & 5.30 & 14.39 & 4.96 & \multirow{3}{*}{14.83} & \multirow{3}{*}{5.18} \\ \cline{2-6}
                                             & 2 & 14.43 & 4.99 & 14.20 & 4.96 & &  \\ \cline{2-6}
                                             & 3 & 14.15 & 4.92 & \textbf{14.06} & \textbf{4.85} & &  \\ \hline
    \end{tabular}
\end{table}

\begin{figure}[t]
    \centering
    \includegraphics[width=0.75\linewidth]{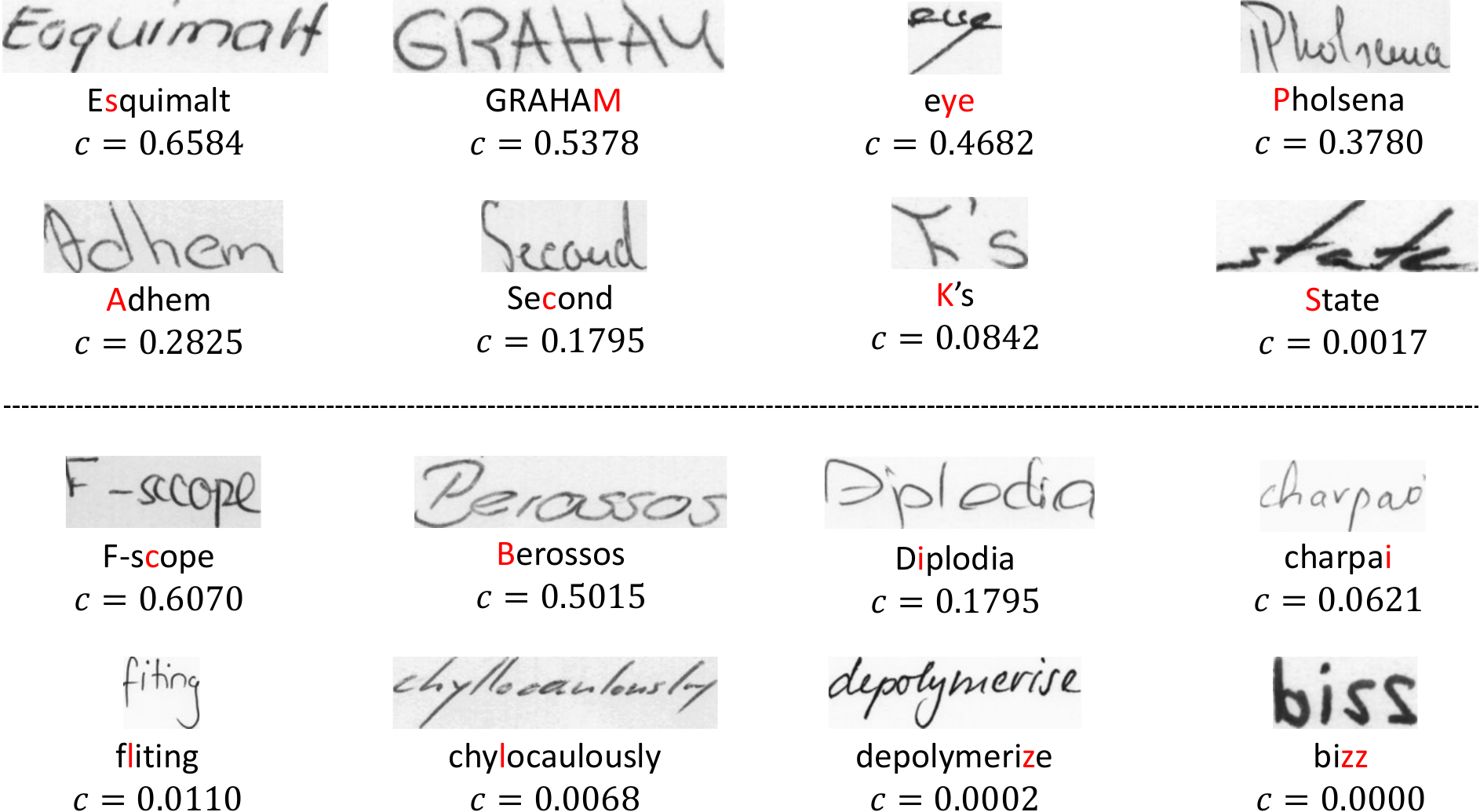}
    \caption{Samples of synthetic handwritten images in (top) Synth-IAM-Words and (bottom) Synth-EN-Words sets that are removed using progressive data filtering. Wrongly generated characters are highlighted in red. Confidence of correctness scores are listed below.  }
    \label{fig:removed-samples}
\end{figure}

To evaluate the effect of the progressive data filtering strategy, we set the number of data filtering rounds $N=3$, and try $\tau=\{ 1.0,~0.7 \}$. We conduct experiments on three synthetic dataset and results are listed in Table \ref{table:wer-cer-filt}. We also list the baseline results when all synthetic samples are added to the training set. The performances of OCR models improve with more progressive data filtering rounds. We also try to use an additional 4th round, but no further performance improvements are observed. Compared with $\tau=1.0$, better performances are achieved using $\tau=0.7$. This implies that $\tau=0.7$ achieves a better tradeoff between numbers of high-quality and noisy samples.
Compared with using all generated samples, progressive data filtering achieves slightly better WER and worse CER on Synth-IAM-Words. After 3 rounds, about $90\%$ and $91\%$ of samples in Synth-IAM-Words are added to the training set with $\tau=1.0$ and $\tau=0.7$, respectively. On Synth-EN-Words and Synth-EN-Words-WI datasets, progressive data filtering achieves much better results than using all generated samples. With progressive data filtering, about $55\%$ and $59\%$ of the data are added to the training set with $\tau=1.0$ and $\tau=0.7$, respectively. 
Fig. \ref{fig:removed-samples} shows samples of synthetic handwritten images that are removed using progressive data filtering. The generated images can contain errors, such as missing or repeating certain characters, or failing to distinguish some easily confused characters. The proposed data filtering strategy can successfully remove these error samples.
These results show that DDPM-generated samples of unseen words are much noisier than samples of seen words, and the progressive data filtering strategy is helpful to remove noisy samples and achieve better OCR performance.

\subsection{Comparison with previous methods}

\begin{figure}[t]
    \centering
    \includegraphics[width=0.65\linewidth]{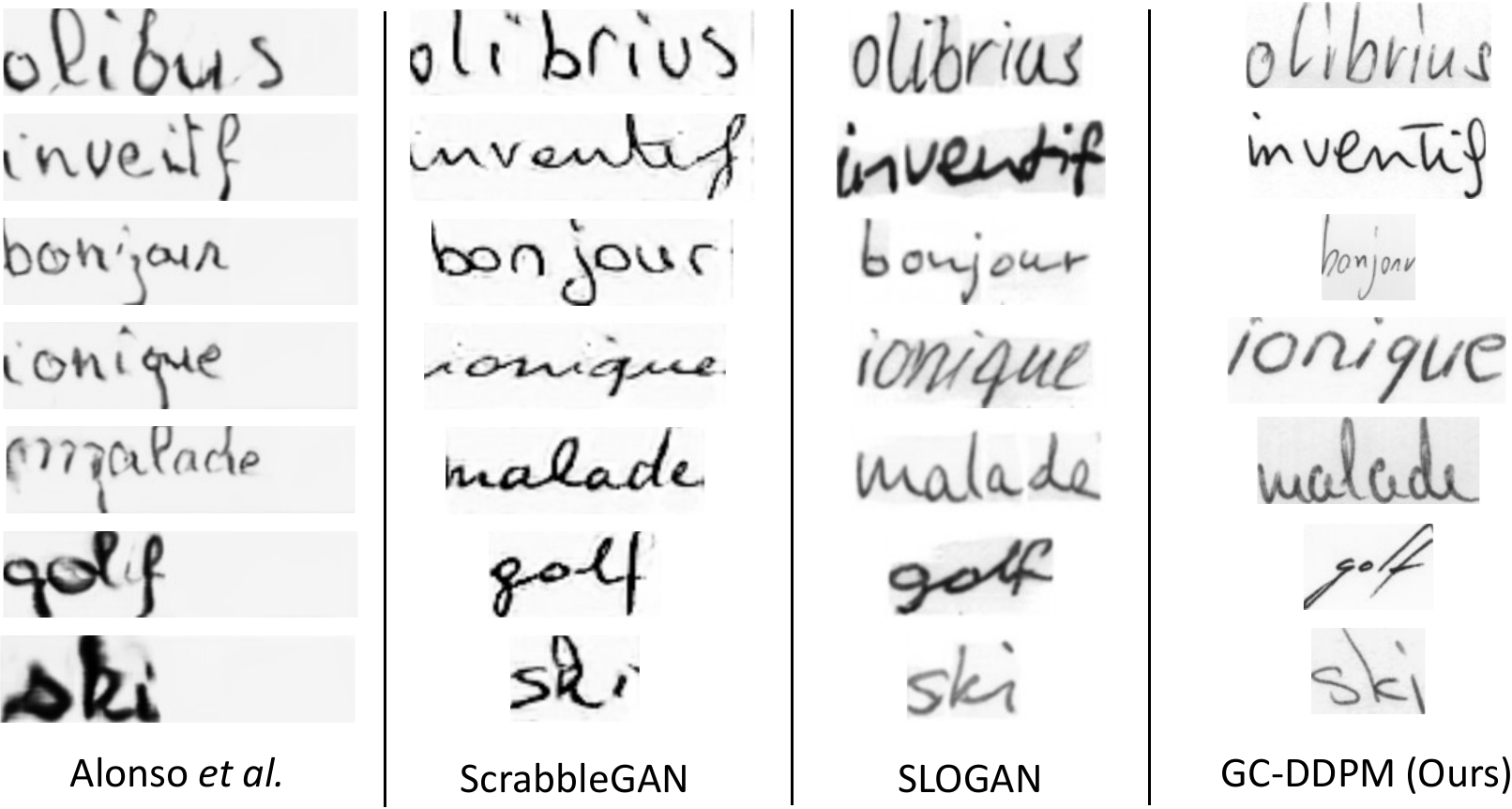}
    \caption{Visual comparisons with Alonso \textit{et al.} \cite{Alonso-ICDAR-2019}, ScrabbleGAN \cite{ScrabbleGAN-2020} and SLOGAN \cite{SLOGAN-2022}. 
    The words generated from top to bottoms are: olibrius, inventif, bonjour, ionique, malade, golf, ski. The writer IDs of our generated samples are 135, 111, 011, 023, 001, 002, 027, respectively.}
		\label{fig:samples-compare}
\end{figure}

\begin{table}[t]
    \centering
    \caption{Comparison with previous methods on IAM word testing set. No lexicons and language models are applied.}
    \label{table:wer-cer-final}
    \begin{tabular}{|l|c|c|c|c|}
        \hline
        Method             & Synthetic data & WER(\%) & CER(\%)  \\
        \hline \hline
        Kang \textit{et al.} \cite{Kang-ICFHR-2020}  & No & 16.39 & 6.43 \\ \hline
        Learn to Augment \cite{Luo-CVPR-2020} + AFDM \cite{Bhunia-CVPR-2019}  &      No & 13.35 & 5.13 \\ \hline
        
        SLOGAN (Baseline) \cite{SLOGAN-2022} &  No & 19.12 & 7.39  \\ \hline \hline

        Dutta \textit{et al.} \cite{Dutta-ICFHR-2018} &  Font-based & 12.61 & 4.88 \\ \hline
        Kang \textit{et al.} \cite{Kang-WACV-2020}    &  Font-based & 17.26 & 6.75 \\ \hline
        SLOGAN \cite{SLOGAN-2022}                     &  GAN-based      & 14.97 & 5.95 \\ \hline
        SLOGAN \cite{SLOGAN-2022} + Learn to Augment \cite{Luo-CVPR-2020}  &  GAN-based      & 12.90 & 4.94 \\ \hline \hline

        Ours (Real)                                &    No                           & 19.47  & 7.27 \\ \hline
        Ours (Real + filtered synthetic data)      &  DDPM-based   & \textbf{10.72} & \textbf{3.75} \\ \hline
        Ours (Filtered synthetic data only)        &   DDPM-based & 11.55 & 4.07 \\ \hline
        
    \end{tabular}
\end{table}

\begin{table}[t]
    \centering
    \caption{Comparison of FID on out-of-vocabulary word images.}
    \label{table:oov-exp}
    \begin{tabular}{|l|c|c|c|}
        \hline
        Method & GANwriting \cite{GANwriting-2020} & SLOGAN  \cite{SLOGAN-2022} & GC-DDPM (ours) \\
        \hline \hline
        FID  & 125.87  & 97.81 & \textbf{86.93}  \\ \hline
    \end{tabular}
\end{table}

Fig. \ref{fig:samples-compare} shows visual comparisons with previous methods. Our GC-DDPM approach can generate photo-realistic handwritten images with fewer artifacts. In ScrabbleGAN \cite{ScrabbleGAN-2020} and SLOGAN \cite{SLOGAN-2022}, the width of the generated image is determined by the length of input text or the width of glyph conditional image. Our approach is able to generate images with variable widths according to the text content and writer style. GC-DDPM can successfully mimic the unique style of the conditional writer styles. For example, writer 111 usually writes ``t'' similarly with ``T'', and the generated stroke of ``t'' in ``inventif'' is also similar to ``T''.

Besides visual comparison, we also compare the synthetic data quality using FID metric. Following GANwriting \cite{GANwriting-2020} and SLOGAN \cite{SLOGAN-2022}, we generate 400  unique out-of-vocabulary (OOV) words and calculate an averaged FID \cite{FID} score of each handwriting style \footnote{According to the authors of GANwriting, the exact list of 400 OOV words is no longer available. Therefore, we follow their advice to build our own OOV word list.}. As shown in Table \ref{table:oov-exp}, we achieve an FID score of 86.93 which is better than that of both GANwriting and SLOGAN.

Since the goal of our approach is to generate handwritten images to augment the training set for handwritten OCR, we compare our approach with other synthetic data augmented OCR systems on the IAM word benchmark. To push the OCR performance to limit, we add all samples in Synth-IAM-Words, filtered Synth-EN-Words and Synth-EN-Words using $\tau=0.7$ to augment the IAM training data. As shown in Table \ref{table:wer-cer-final}, we achieve a $10.72\%$ WER and a $3.75\%$ CER, which are much better than previous methods using font rendered or GAN-based synthetic images. Compared with using real IAM training data alone, we achieve a relative $45\%$ WERR and a relative $48\%$ CERR. We also conduct an experiment of using only filtered synthetic data, and achieve a $11.55\%$ WER and a $4.07\%$ CER, which are significantly better than using IAM real training data only.

\subsection{Experiments on IAM text line dataset} \label{sec:exp-line}

\begin{figure}[t]
    \centering
    \includegraphics[width=0.6\linewidth]{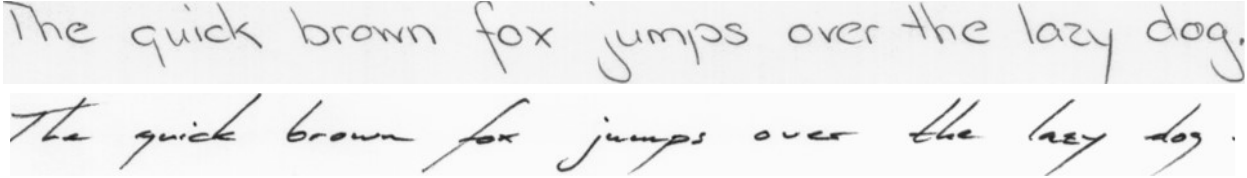}
    \caption{Samples of synthetic handwritten text line images of a sentence: ``The quick brown fox jumps over the lazy dog.'' .}
    \label{fig:samples-lines}
\end{figure}

\begin{table}[t]
    \centering
    \caption{Comparison with previous methods on IAM text line testing set. No lexicons and language models are applied.}
    \label{table:wer-cer-line-final}
    \begin{tabular}{|l|c|c|c|}
        \hline
        Method           & Synthetic data & WER(\%) & CER(\%)  \\
        \hline \hline
        Puigcerver \cite{Puigcerver-ICDAR-2017} & No & 18.40 & 5.80 \\ \hline
        Michael \textit{et al.} \cite{Wang-ICDAR-2019} & No & - & 5.24 \\ \hline
        Wick \textit{et al.} \cite{HWR-ICDAR-2021-1}   & No & - & 5.67 \\ \hline\hline

        Dutta \textit{et al.} \cite{Dutta-ICFHR-2018} & Font-based & 17.82 & 5.70 \\ \hline
        Barrere \textit{et al.} \cite{Barrere-DAS-2022} & Font-based & 16.31 & 4.76 \\ \hline
        Kang \textit{et al.} \cite{Kang-PR-2022}    & Font-based & 15.45 & 4.67 \\ \hline

        Wick \textit{et al.} (CTC) \cite{Wick-DAS-2022}   & Font-based & 16.85 & 4.99 \\ \hline
        Wick \textit{et al.} \cite{Wick-DAS-2022}   & Font-based & 12.20 & 3.96 \\ \hline
        $\mathrm{TrOCR}_{\mathrm{SMALL}}$ \cite{TROCR-2021} & Font-based & - & 4.22 \\ \hline \hline

        Ours (Real)                                 & No  & 22.11 & 7.05 \\ \hline
        Ours (Real + filtered synthetic data)       & DDPM-based & 13.08 & 4.13 \\ \hline
    \end{tabular}
\end{table}

Finally, we conduct experiments on the IAM text line benchmark. Although the GC-DDPM is trained on images with maximum aspect ratio of 8, it can generate text lines as shown in Fig. \ref{fig:samples-lines}. We use the same CTC-based OCR model as in IAM word experiments, and achieve a $7.05\%$ CER. For synthetic data, besides using the filtered synthetic word dataset, we also synthesize handwritten text line samples using the IAM training line corpus, and filter these samples using progressive data filtering. As shown in Table \ref{table:wer-cer-line-final}, we finally achieve a $4.13\%$ CER, which is slightly better than $\mathrm{TrOCR}_{\mathrm{SMALL}}$. It should be noted that works in \cite{Wick-DAS-2022,TROCR-2021} use advanced sequence-to-sequence framework for OCR. \cite{Wick-DAS-2022} achieves a $4.99\%$ CER with CTC-based model.
$\mathrm{TrOCR}_{\mathrm{SMALL}}$ also leverages pre-trained encoder and decoder with 62M total parameters. We only use a simple CTC-based OCR model with 14M parameters, without using any image pre-processing technique and pre-trained models.

\section{Conclusion} \label{sec:conclude}

In this paper, we investigate GC-DDPM to generate handwritten images to augment training data for handwritten OCR. The proposed GC-DDPM is able to generate photo-realistic handwritten samples with diverse styles and text contents. However, we find that the text contents in synthetic images are not always consistent with the glyph conditional images, especially in images with out-of-vocabulary words. Therefore, we further propose a progressive data filtering method to remove samples with noisy labels. Experiments on both IAM word and text line benchmarks show that the performance of the OCR model trained with augmented DDPM-synthesized samples can perform much better than the one trained on real data only.



%
%
%
\bibliographystyle{splncs04}
\bibliography{reference}

\end{document}